# Neural Optimization Machine: A Neural Network Approach for Optimization

Jie Chen[a], Yongming Liu[b,*]

[a] *Department of Mechanical Engineering, Northwestern University, Evanston, IL 60208, USA, E-mail: jie.chen@northwestern.edu*
[b] *School for Engineering of Matter, Transport, and Energy, Arizona State University, Tempe, AZ 85287, USA, E-mail: Yongming.Liu@asu.edu*

**Abstract:** A novel neural network (NN) approach is proposed for constrained optimization. The proposed method uses a specially designed NN architecture and training/optimization procedure called Neural Optimization Machine (NOM). The objective functions for the NOM are approximated with NN models. The optimization process is conducted by the neural network's built-in backpropagation algorithm. The NOM solves optimization problems by extending the architecture of the NN objective function model. This is achieved by appropriately designing the NOM's structure, activation function, and loss function. The NN objective function can have arbitrary architectures and activation functions. The application of the NOM is not limited to specific optimization problems, e.g., linear and quadratic programming. It is shown that the increase of dimension of design variables does not increase the computational cost significantly. Then, the NOM is extended for multiobjective optimization. Finally, the NOM is tested using numerical optimization problems and applied for the optimal design of processing parameters in additive manufacturing.

## 1. Introduction

Optimization is used in various scientific and engineering applications. Examples contain function, approximation, and regression analysis optimal control, system planning, signal processing, mechanical design (Effati & Nazemi, 2006). One possible and auspicious method for optimization is utilizing neural networks (NNs) due to their inherent massive parallelism (Effati & Nazemi, 2006; Lopez-Garcia, Coronado-Mendoza, & Domínguez-Navarro, 2020), the ability to deal with time-varying parameters (Xia, Feng, & Wang, 2008), the robustness of the computation (X.-S. Zhang, 2013), and a large community for rapid advancement in recent years (Chen, Gao, & Liu, 2022; C. Wu, Wang, & Kim, 2022).

Tank and Hopfield (Tank & Hopfield, 1986) proposed several neural optimization networks by applying circuit theory in optimization using neural networks (X.-S. Zhang, 2013). Dhingra and Rao (Dhingra & Rao, 1992) adapted Hopfield's neural network to solve nonlinear programming problems with/without constraints. The solutions from the neural network approach agree well with those calculated by gradient-based search techniques. Wu et al. (A. Wu & Tam, 1999) presented a neural network method for quadratic programming problems by applying the Lagrange multiplier theory. The solutions satisfy the necessary conditions for optimality. The connections in neural networks are designed according to the optimization problems. Tagliarini et al. (Tagliarini, Christ, & Page, 1991) presented a rule for neural network design for optimization problems. Time evolution equations are constructed with equality and inequality constraints. Effati et al. (Effati & Nazemi, 2006) solved linear and quadratic programming problems using recurrent neural networks. The NN approach establishes an energy function and a dynamic system. The solution can be found when the dynamic system approaches its static state. Xia et al. (Xia, et al., 2008) developed a recurrent neural network approach for optimization under nonlinear inequality constraints. That model was developed for convex optimization problems, and is suitable for only a specific category of nonconvex optimization problems.

Another category of applying neural networks in optimization is to use the NN as a surrogate model. Neelakantan et al. (Neelakantan & Pundarikanthan, 2000) trained a NN to approximate the simulation model and applied nonlinear programming methods to find near-optimal policies. Nascimento et al. (C. A. O. Nascimento, Giudici, & Guardani, 2000) used the NN to replace the equations of optimization problems. First, a grid search is carried out in the region of interest. Then, the solutions violating the constraints are excluded. This method allows one to identify multiple optima. Darvishvand et al. (Darvishvand, Kamkari, & Kowsary, 2018) trained NNs to construct the objective functions and then used Genetic Algorithm to find the optimal design variables. Villarrubia et al. (Villarrubia, De Paz, Chamoso, & la Prieta, 2018) used NNs to solve problems when the linear programming or Lagrange multiplier is not applicable. First, the NN is used for objective function approximation. Then Lagrange method is used to solve optimization problems. Jeon et al. (Jeon, Lee, & Choi, 2019) proposed a double-loop process for training and optimizing the NN objective function. The outer process is to optimize the NN weights. The inner process aims to optimize input variables while fixing the NN weights. Chandrasekhar et al. (Chandrasekhar & Suresh, 2021) used the NN as the density function for topology optimization. The inputs are location coordinates, and the outputs are density values. The density field is optimized by relying on the NN's backpropagation and a finite element solver.

This paper proposes a neural network approach for optimizing neural network surrogate models with and without constraints. The method is called Neural Optimization Machine (NOM). The NOM has the following features. The objective functions for the NOM are NN models. The optimization process is conducted by the neural network's built-in backpropagation algorithm. The NOM solves optimization problems by extending the architecture of the NN objective function model. This is achieved by appropriately designing the NOM's structure, activation function, and loss function. The proposed NOM has the following benefits. First, the NOM is very flexible, and the NN objective function can have arbitrary architectures and activation functions. Second, the NOM is not limited to specific optimization problems, e.g., linear and quadratic programming. Third, multiple local minima can be found, which provides the potential for finding the global minimum. Fourth, compared with current heuristics optimization techniques, e.g., Particle Swarm Optimization and Genetic Algorithm, the increase of dimension of design variables does not increase the computational cost significantly. Fifth, the NOM can be easily extended and adapted for multiobjective optimization.

The remainder of the paper is organized as follows. First, a brief introduction to neural networks is provided. Following this, the methodology of the Neural Optimization Machine is presented. The construction of the NOM for unconstrained and constrained optimization is described. The NOM is then extended for multiobjective optimization. Next, the NOM is tested using numerical optimization problems and applied for the design in additive manufacturing. Next, a discussion is given regarding the NOM finding multiple local minima. Finally, the conclusions are drawn according to the current study.

## 2. Neural Optimization Machine

### 2.1 A brief introduction of neural networks

Fig. 1 shows an illustrative example for explaining the main concepts of neural networks (Chen & Liu, 2020a, 2021a, 2021b; Laddach, Łangowski, Rutkowski, & Puchalski, 2022). There are three layers in this example. The first and the last layer are Input and Output Layer, respectively. The layer in between is Hidden layer. The nodes in each layer are called neurons. There are two and one neurons at the first and last layer in this example. Thus, it is a two-dimensional problem with one output. The neural network in this example is called feedforward neural network since information flows layer by layer from the Input layer to Output Layer.

The mathematical calculation in each neuron can be expressed as follows (Sattarifar & Nestorović, 2022):

$$z_k^{(l)} = b_k^{(l-1)} + \sum_{j=1}^{p_{l-1}} w_{kj}^{(l-1)} a_j^{(l-1)}, \quad l = 2,...,L, \tag{1}$$

$$a_k^{(l)} = G_k^{(l)}\left(z_k^{(l)}\right), \quad k = 1, 2, ..., p_l, \tag{2}$$

where $l$ is the layer index. $l$ is 1 and $L$ for Input and Output Layer, respectively. The number of neurons at the $l$th is $p_l$. At the first layer (input layer),

$$a_k^{(1)} = x_k, \quad k = 1,...,p_1. \tag{3}$$

The $k$th neuron at layer $l$ has activation function $G_k^{(l)}(\cdot)$. In neural networks, the coefficients and intercepts in Eq. (1) are named weights and biases, respectively. During the training of the neural network, weights and biases are updated while minimizing the loss function. The loss function is the objective function in the training process (Ketkar, 2017; R. G. Nascimento, Fricke, & Viana, 2020). A common optimization technique in neural networks is stochastic gradient descent. To alleviate the computational burden, training the neural network is carried out though mini-batches of the total training data. In each epoch, the training data are used in a mini-batch manner until all the data are used. A number of epochs are needed to train a neural network.

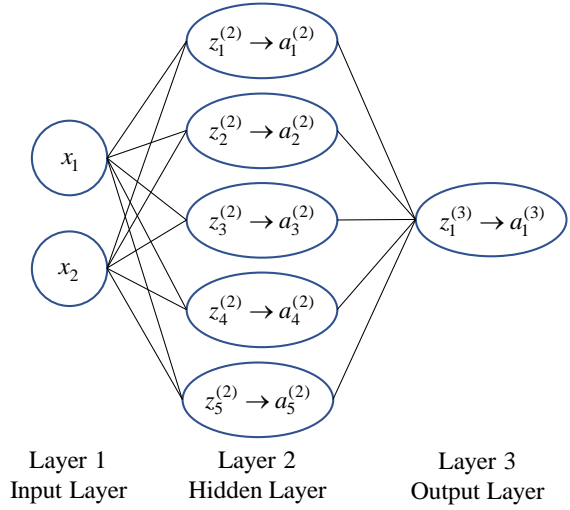

Fig. 1 An illustrative example of a single hidden layer neural network.

## 2.2 Neural Optimization Machine (NOM)

Suppose a neural network model *NN* (**X**) has been trained. The *NN* (**X**) is then used as the objective function in optimization. The goal is to find the set of inputs **X**, which minimize the outputs of *NN* (**X**) under inequality and equality constraints. The problem of the constrained optimization is formulated as

Minimize
$$f(\mathbf{X}) = NN(\mathbf{X})$$
subject to
$$g_p(\mathbf{X}) \leq 0, \quad p = 1, ..., P$$
$$h_q(\mathbf{X}) = 0, \quad q = 1, ..., Q$$
(4)

where **X** is the input vector, *NN* (**X**) is the NN model used as the objective function, and *g* and *h* are inequality and equality constraints, respectively. A simple neural network model *NN* (**X**) in Fig. 2 is used as the NN objective function for illustration, which is the same as the architecture in Fig. 1.

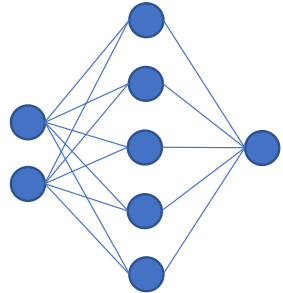

Fig. 2 A simple neural network.

The key idea of the developed Neural Optimization Machine (NOM) is to solve the optimization problem in Eq. (4) using the neural networks' built-in backpropagation algorithm by properly designing the NN architecture. On the one hand, in the backpropagation algorithm, the basic method is stochastic gradient descent. It computes the gradients of the loss function with respect to the weights and biases. The weights and biases are then updated by the gradient information (Z. Zhang, Li, & Lu, 2021). At the end of the training, a set of local optimal weights and biases are obtained. On the other hand, the gradient descent method can also be used to solve the optimization problem in Eq. (4). It requires computing the gradient of the outputs of the NN objective function with respect to its inputs. Then, the question is, *can we transform the problem of calculating the gradient of NN outputs with respect to inputs to the problem of calculating the gradient of the NN loss function with respect to weights and biases*? If this can be achieved, another question is *how to consider the constraints in Eq. (4)*. The NOM is developed to solve those two issues.

### 2.2.1 NOM for unconstrained optimization

To illustrate the basic components of the Neural Optimization Machine, we will first consider the unconstrained optimization problem, i.e., the problem without the constraints in Eq. (4). The NOM architecture in Fig. 3 is designed to answer the first question, that is, transforming the problem of calculating the gradient of NN outputs with respect to inputs to the problem of calculating the gradient of the NN loss function with respect to weights and biases. The subpart of the NOM architecture shown in the dashed line box in Fig. 3 is the trained NN objective function to be optimized, as shown in Fig. 2. It is called NN objective function to differentiate it from the NOM. A new layer, called starting point layer, is added before the input layer of the NN objective function. There are two starting point neurons in this layer for this example, as shown in grey. Each neuron in the starting point layer is connected to one of the neurons in the input layer. No activation functions are applied in this layer. The values input to the NN objective function are controlled by the starting points and weights and biases between the starting point layer and the input layer. Following the formulation in Eqs. (1) - (3), the inputs to the NN objective function are calculated by

$$z_k^{(1)} = b_k^{(s)} + w_{kk}^{(s)} a_k^{(s)} \quad k = 1, 2, ..., p_1, \tag{5}$$

$$x_k = a_k^{(1)} = z_k^{(1)}, \tag{6}$$

where the superscript (*s*) and (1) indicate the starting point layer of the NOM and the input layer of the NN objective function, respectively, and $p_1$ is the input dimension of the NN objective function. The optimal $x_k$, $k = 1, 2, ..., p_1$, are desired to minimize the NN objective function. They are obtained from Eq. (6) by training the NOM.

To achieve this goal, we customize the loss function of NOM and constrain the NN weights and biases. The loss function of the NOM is the output of the NOM:

$$Loss(\text{NOM}) = \text{NOM output} = NN(\mathbf{X}), \tag{7}$$

In this case, the output of the NOM is the same as the output of the NN objective function. That is, the NOM is trained to minimize the output of the NN objective function. In Fig. 3, different colors of the connections between neurons mean whether the weights and biases are fixed (orange) or trainable (blue). The weights and biases are fixed for the NN objective function in the dashed line box and are trainable for the connection between the starting point layer and the input layer. This is because when training the NOM, the original NN model (i.e., NN objective function) should be kept unchanged while the weights and biases between the starting point layer and the input layer are updated to find the optimal solution to minimize the NOM.

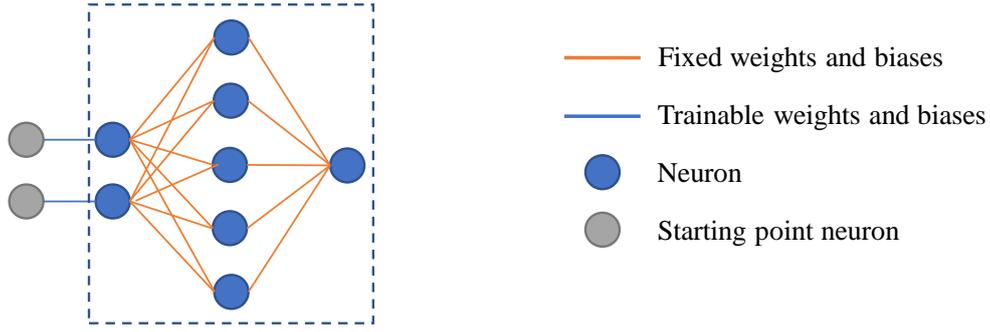

Fig. 3 Neural Optimization Machine for unconstrained optimization.

Only one training data point is used to train the NOM. The data point is referred to as the starting point of the stochastic gradient descent algorithm in optimizing the NOM. This data can also be interpreted as the training data for the input of the NOM (i.e., the starting point is the input to the starting point layer of the NOM). There is no training data for the output of NOM since the NOM is not trained in a supervised way.

A general drawback of NN training is the convergence to local minima. As a result, the NOM will produce a locally optimal solution. The following two operations are adopted to mitigate this issue and possibly find the global minimum. First, a grid search is conducted on the region of interest to generate multiple good starting points. The constraints can then be evaluated if needed to exclude the points with constraint violations (C. A. O. Nascimento, et al., 2000). Next, each good starting point is used as the training data of the NOM to find the corresponding optimal solution to the NN objective function. Second, before training the NOM, random initial values are assigned to the weights between the starting point layer and the input layer. This is a default feature of the optimizer in training the NN. This operation adds some randomness to the initial search directions. The effects of the above operations are further discussed in Section 4.

### 2.2.2   NOM for constrained optimization

The above description of the NOM shown in Fig. 3 is for unconstrained optimization. On the basis of the NOM for unconstrained optimization, the NOM is further developed and shown in Fig. 4 for answering the second question, i.e., how to consider the constraints. Compared with the NOM in Fig. 3, a constraint layer is added. The neurons in this layer are called constraint neurons, as shown in green. The number of constraint neurons is equal to the number of constraints. Following Eqs. (1) and (2), the calculation from the input layer to the constraint layer can be expressed as

$$z_k^{(c)} = g_k(\mathbf{X}), \quad k = 1,...,P, \tag{8}$$

for inequality constraints, and

$$z_k^{(c)} = h_k(\mathbf{X}), \quad k = P+1,...,P+Q, \tag{9}$$

for equality constraints, and then

$$a_k^{(c)} = G_k^{(c)}\left(z_k^{(c)}\right), \quad k = 1,2,...,P+Q, \tag{10}$$

where the superscript ($c$) indicates the constraint layer. The adopted activation function $G^{(c)}$ for the constraint layer is modified from the Rectified Linear Unit (ReLU) activation function, which is expressed as

$$\text{ReLU}(z) = \max(0, z). \tag{11}$$

For the inequality constraints shown in problem (4), the activation function used for the constraint neuron is

$$G^{(c)}(z) = c \cdot \text{ReLU}(z), \tag{12}$$

and for the equality constraints,

$$G^{(c)}(z) = c \cdot \left[ \text{ReLU}(-z) + \text{ReLU}(z) \right], \tag{13}$$

where $c$ is a large number as the penalty parameter. The plots of ReLU (Eq. (11)) and two modified ReLU activation functions (Eqs. (12) and (13)) with $c = 10$ are plotted in Fig. 5. The physics meaning of using the above activation functions is that when the constraints are satisfied, the outputs of the constraint neurons are zero, and when the constraints are violated, the outputs of the constraint neutrons are large values.

Another difference between the NOM architectures in Fig. 3 and Fig. 4 is the output of the NOM. In Fig. 3, the output of the NOM is the same as the output of the NN objective function. In Fig. 4, the NOM output is the sum of the NN objective function $NN(\mathbf{X})$ and the outputs of all constraint neurons. The loss function of the NOM for constrained optimization is still the output of the NOM, i.e.,

$$Loss(\text{NOM}) = \text{NOM output} = NN(\mathbf{X}) + \text{outputs of constraint neurons}, \tag{14}$$

The meaning of Eq. (14) is that when the constraints are violated (i.e., the inputs of the NN objective function are infeasible), a large penalty value is added to the loss function. The NOM is trained to minimize the loss function Eq. (14), which brings the inputs to the NN objective function to the feasible domain by adjusting the weights and biases between the starting point layer and the input layer.

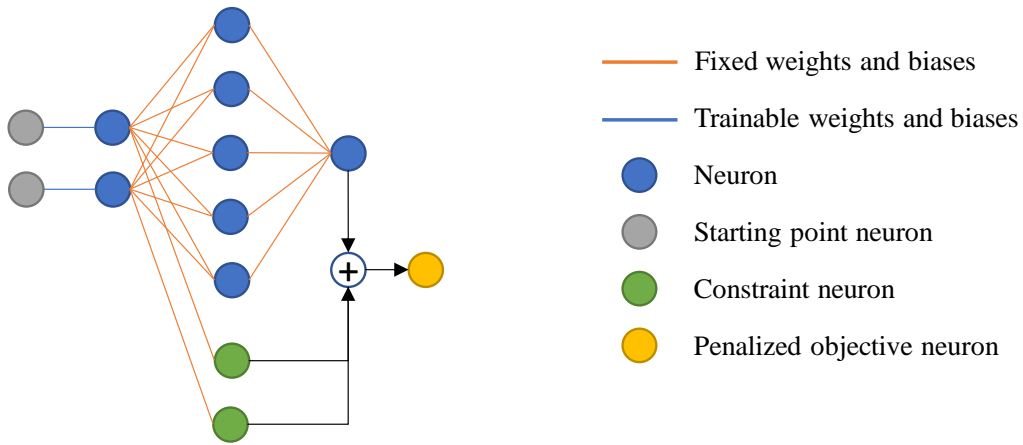

Fig. 4 Neural Optimization Machine for constrained optimization.

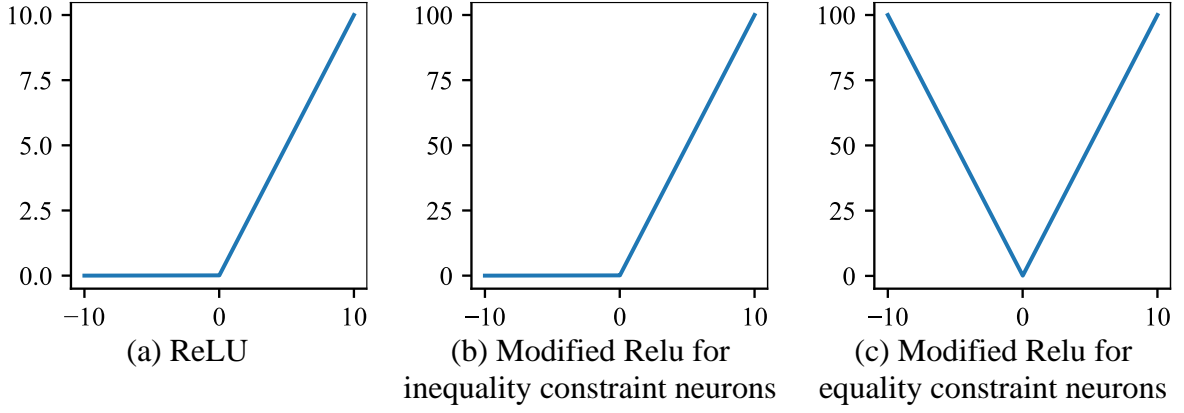

Fig. 5 ReLU and modified ReLU activation functions.

### 2.3 NOM for multiobjective optimization

The proposed Neural Optimization Machine is a flexible tool. It can be developed to solve other types of optimization problems involving NNs as the objective functions. This section shows how the NOM can be developed for multiobjective optimization for NNs.

The multiobjective optimization problem is formulated as

$$\begin{aligned}\text{Minimize} \\ \mathbf{F}(\mathbf{X}) &= \{f_1(\mathbf{X}), f_2(\mathbf{X})..., f_M(\mathbf{X})\} \\ &= \{NN_1(\mathbf{X}), NN_2(\mathbf{X})..., NN_M(\mathbf{X})\} \\ \text{subject to} \\ g_p(\mathbf{X}) &\leq 0, \quad p = 1,..., P \\ h_q(\mathbf{X}) &= 0, \quad q = 1,..., Q \end{aligned} \quad (15)$$

where the $M$ objective functions are $M$ NN models. Usually, there is no $\mathbf{X}$ that can minimize all objective functions simultaneously (Rao, 2019). Infinite solutions can be obtained for different combinations of good performance of objectives. Those solutions are called Pareto optimum solutions (Mack, Goel, Shyy, & Haftka, 2007; Martínez-Iranzo, Herrero, Sanchis, Blasco, & García-Nieto, 2009; Sundaram, 2022). Pareto optimal is a feasible solution X that satisfy the following condition: we cannot find another feasible solution $\mathbf{Y}$ that $f_i(\mathbf{Y}) \leq f_i(\mathbf{X})$ for all objectives ($i = 1, 2, ..., M$) and $f_j(\mathbf{Y}) < f_j(\mathbf{X})$ for at least one objective function. (Rao, 2019). The goal of multiobjective optimization is to find Pareto optimum solution set.

Among several methods developed for multiobjective optimization, the bounded objective function method is used, which is suitable for the proposed Neural Optimization Machine. Consider the maximum and minimum acceptable values for objective function $f_i$ are $U^{(i)}$ and $L^{(i)}$ ($i = 1, 2, …, M$), respectively. We minimize the most important (say, the $r$th) objective function (Rao, 2019):

$$\begin{aligned}\text{Minimize} \\ f_r(\mathbf{X}) &= NN_r(\mathbf{X}) \\ \text{subject to} \end{aligned} \quad (16)$$

$$g_p(\mathbf{X}) \leq 0, \quad p = 1,...,P$$
$$h_q(\mathbf{X}) = 0, \quad q = 1,...,Q$$
$$L^{(i)} \leq f_i(\mathbf{X}) = NN_i(\mathbf{X}) \leq U^{(i)}, \quad i = 1,...,M, i \neq r$$

$L^{(i)}$ can be discarded if the goal is not to achieve a solution falling within a range of values each objective. In that case, $U^{(i)}$ can be varied systematically to produce Pareto optimal solution set (Arora, 2004). That is adopted in this paper. A general guideline for selecting $U^{(i)}$ is (Carmichael, 1980)

$$f_i(\mathbf{X}_i^*) \leq U^{(i)} \leq f_r(\mathbf{X}_i^*). \tag{17}$$

More discussion on selecting $U^{(i)}$ can be found in (Cohon, 2004; Stadler, 1988).

A numerical example with two NN objective functions is used to illustrate the proposed NOM in multiobjective optimization using the bounded objective function method. The NOM for multiobjective optimization is shown in Fig. 6. The subparts of the NOM in the two dashed line boxes are the two NN objective functions, $NN_1$ and $NN_2$, respectively. The architecture of each NN objective function is shown in Fig. 2. Compared with the NOM in Fig. 4, another NN objective function is added in Fig. 6. Both NN objective functions share the same input layer in this example. According to the bounded objective function method, $NN_1$ is used as the objective function, and $NN_2$ is used as the constraint. The output neuron of $NN_2$ is replaced by the constraint neuron. The upper bound, $U^{(2)}$, changes according to Eq. (17). For each $U^{(2)}$, a Pareto optimal solution is obtained using the NOM. Pareto optimal solutions can be obtained by varying the $U^{(2)}$.

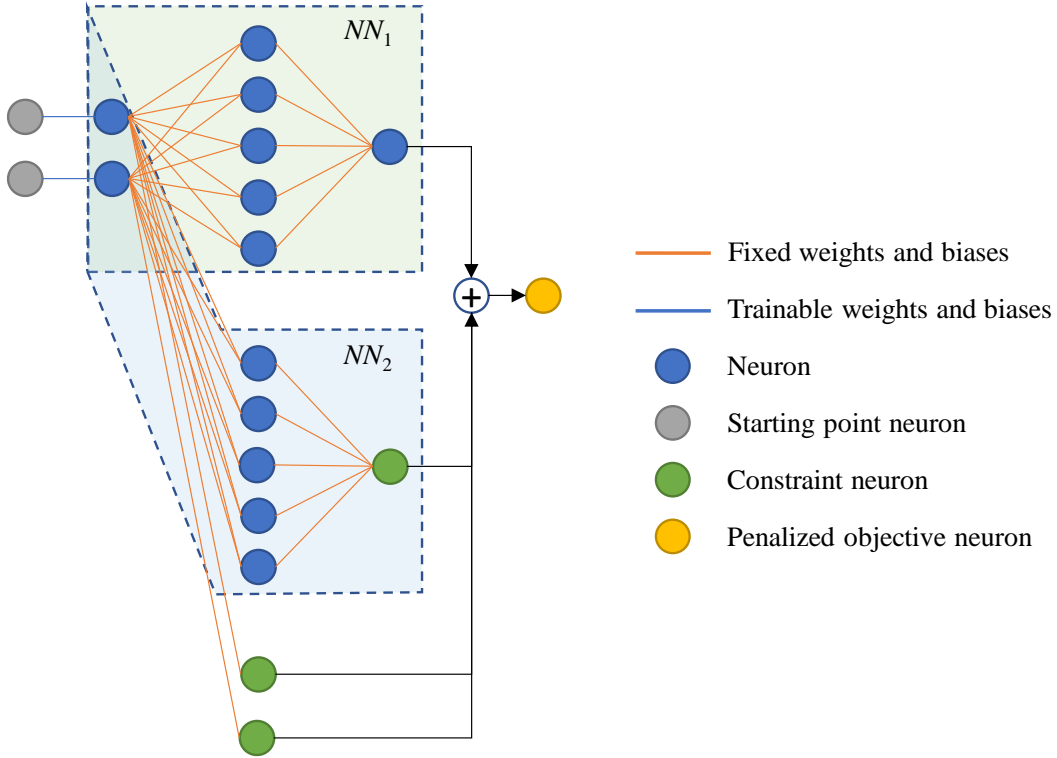

Fig. 6 Neural Optimization Machine for multiobjective optimization.

## 3. Experiments and applications

In this section, the proposed Neural Optimization Machine (NOM) is tested for numerical problems of constrained optimization and multiobjective optimization. Following that, the NOM is applied to design processing parameters in additive manufacturing.

### 3.1 Constrained optimization problems

In this subsection, three optimization problems are analyzed. The results from the NOM are compared with those from Nelder Mead, Generic algorithm, Differential Evolution, and Particle Swarm Optimization.

The data used to train the neural network (NN) objective function are generated inside the domain defined by the constraints on individual variables. 10,000 data are generated evenly for each problem. The hyperparameters for training the NN objective functions and the NOM are identical for all three problems and are listed in Table 1 and Table 2, respectively.

Table 1 The hyperparameters for training the NN objective functions.

| | |
|---|---|
| Number of hidden layers | 1 |
| Number of hidden neurons | 20 |
| The activation function of hidden layers | Hyperbolic tangent (tanh) |
| Epochs | 500 |
| Minibatch size | 10 |
| Learning rate | 0.01 |

Table 2 The hyperparameters for the NOM.

| | |
|---|---|
| Number of starting points | 5 |
| Penalty parameters | 10 |
| Epochs | 2000 |
| Minibatch size | 1 |
| Learning rate | 0.01 |

The first problem (Problem 1) has a cubic objective function with quadratic constraints:

Minimize

$$f(x_1, x_2) = \frac{1}{1000}\left[(x_1 - 10)^3 + (x_2 - 10)^3\right]$$

subject to

$$-(x_1 - 5)^2 - (x_2 - 5)^2 + 100 \leq 0 \quad (18)$$
$$(x_1 - 6)^2 - (x_2 - 5)^2 - 100 \leq 0$$
$$13 \leq x_1 \leq 20$$
$$0 \leq x_2 \leq 20$$

The second problem (Problem 2) have a trigonometric objective function with linear and quadratic constraints:

Minimize $\quad(19)$

$$f(x_1, x_2) = -10\cos(x_1 x_2) + x_1 x_2/10 + 10(x_1 + x_2)\sin(x_1 + x_2)$$

subject to

$$x_1 - x_2 \geq 0.5$$
$$x_1 x_2 \leq 15$$
$$0 \leq x_1 \leq 1.5$$
$$-1 \leq x_2 \leq 1$$

The third problem (Problem 3) is to optimize a trigonometric objective function with quadratic constraints:

Minimize

$$f(x_1, x_2) = \frac{\sin^3(2x_1)\sin(2x_2)}{x_1^3(x_1 + x_2)}$$

subject to

$$x_1^2 - x_2 + 1 \leq 0$$
$$(x_1 - 2)^2 - x_2 + 1 \leq 0$$
$$0.1 \leq x_1 \leq 1$$
$$0.1 \leq x_2 \leq 7$$

(20)

The plots of the three objective functions are shown in Fig. 7. There are zero, one, and two local minima in the defined field for Problem 1, 2, and 3, respectively. The results of the NOM are shown in Table 3. The results are compared with those of Nelder Mead, Genetic Algorithm, Differential Evolution, and Particle Swarm Optimization, which are also shown in Table 3. The results from the NOM are almost identical to those of other optimization algorithms. The running time is also shown in Table 3. The direct search method (Nelder Mead) uses the shorted time. The NOM uses less computational time than heuristics algorithms (Genetic Algorithm, Differential Evolution, Particle Swarm Optimization).

Table 3 Results of different optimization algorithms.

|  |  | Nelder Mead | Genetic Algorithm | Differential Evolution | Particle Swarm Optimization | Neural Optimization Machine |
|---|---|---|---|---|---|---|
| Problem 1 | $x_1$ | 13.660 | 13.660 | 13.660 | 13.694 | 13.680 |
|  | $x_2$ | 0.000 | 0.000 | 0.000 | 0.000 | 0.001 |
|  | $f(x_1, x_2)$ | -7.950 | -7.950 | -7.950 | -7.949 | -7.948 |
|  | Time (s) | 3.2 | 55.2 | 137.4 | 36.0 | 25.7 |
| Problem 2 | $x_1$ | 0.258 | 0.257 | 0.256 | 0.256 | 0.257 |
|  | $x_2$ | -0.241 | -0.242 | -0.243 | -0.243 | -0.243 |
|  | $f(x_1, x_2)$ | -9.983 | -9.984 | -9.984 | -9.984 | -9.984 |
|  | Time (s) | 3.0 | 54.6 | 139.7 | 35.3 | 20.0 |
| Problem 3 | $x_1$ | 0.100 | 0.100 | 0.100 | 0.100 | 0.101 |
|  | $x_2$ | 5.463 | 5.464 | 5.464 | 5.464 | 5.464 |
|  | $f(x_1, x_2)$ | -1.406 | -1.406 | -1.406 | -1.406 | -1.405 |
|  | Time (s) | 3.0 | 57.5 | 143.5 | 34.8 | 21.8 |

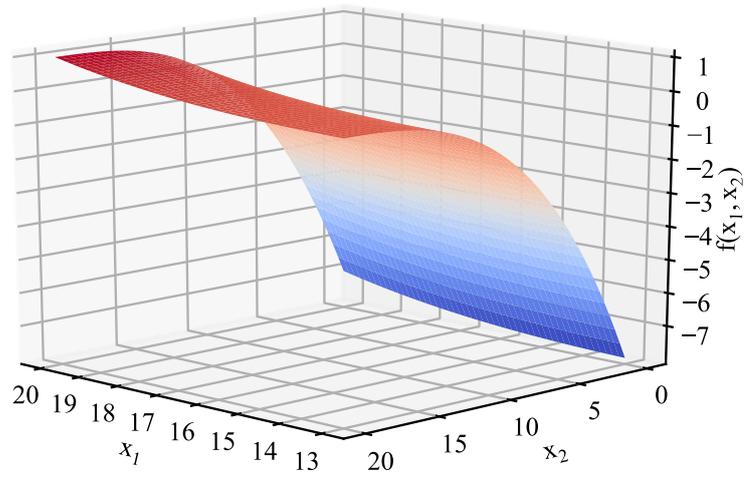

(a) Problem 1.

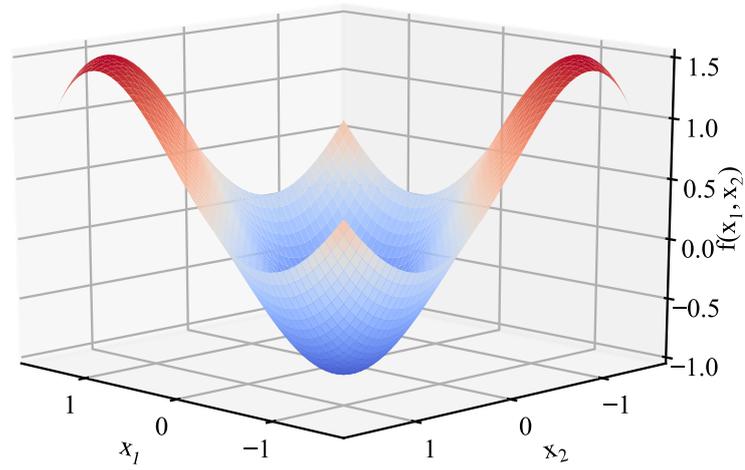

(b) Problem 2.

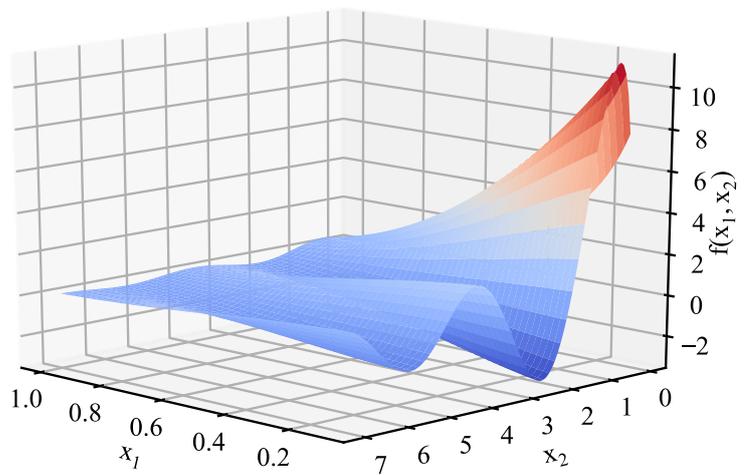

## 3.2 Multiobjective optimization problem

The NOM is tested for multiobjective optimization following the methodology presented in Section 2.3.

The multiobjective optimization problem is shown in Eqs. (21). This is a problem with two quadratic objective functions and linear constraints. The plots of the objective functions are shown in Fig. 8.

Minimize
$$f_1 = (x_1 - 3)^2 + (x_2 - 7)^2$$
$$f_2 = (x_1 - 9)^2 + (x_2 - 8)^2$$

subject to
$$70 - 4x_2 - 8x_1 \leq 0$$
$$-2.5x_2 + 3x_1 \leq 0$$
$$-6.8 + x_1 \leq 0$$
$$0 \leq x_1 \leq 10$$
$$5 \leq x_2 \leq 15$$
(21)

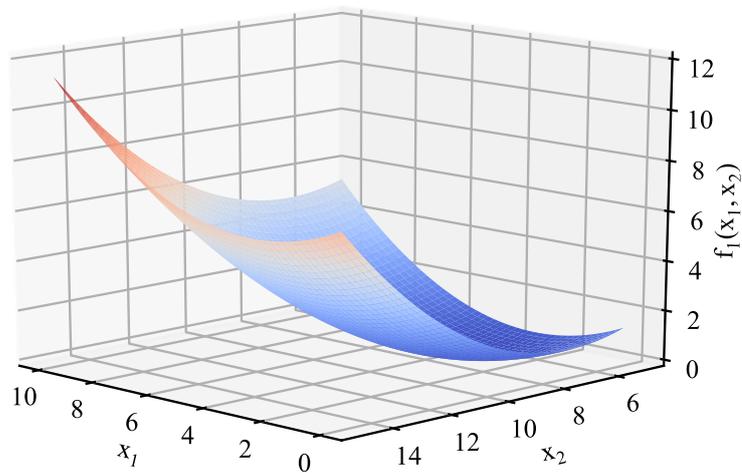

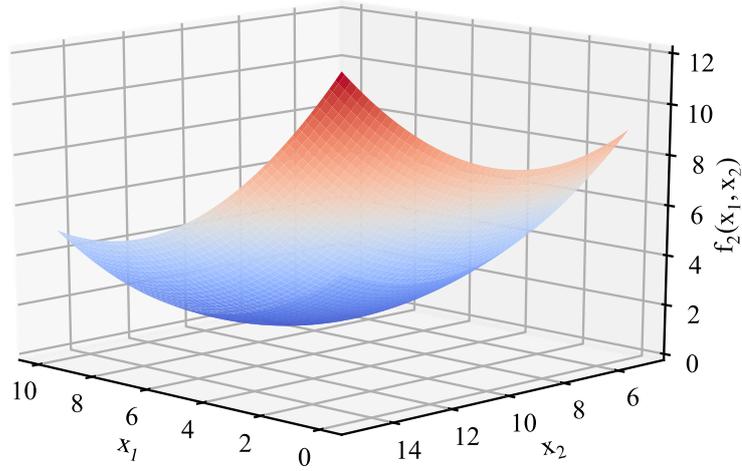

Fig. 8 Plots of the objective functions of the multiobjective optimization problem.

The architecture shown in Fig. 6 is used for this problem. According to the bounded objective function method setup in Eq. (16), $f_1$ is used as the single objective function, and $f_2$ is used as one of the constraints. The optimal solution $\mathbf{X}_2^*$ of $f_2$ is $x_1 = 6.784$ and $x_2 = 8.309$. The outcomes of $f_2$ and $f_1$ at $\mathbf{X}_2^*$, $f_2(\mathbf{X}_2^*)$, and $f_1(\mathbf{X}_2^*)$, are 0.501 and 1.611, respectively. According to Expression (17), $f_2(\mathbf{X}_2^*)$ and $f_1(\mathbf{X}_2^*)$ are used as the lower and upper limits for varying the constraint of $f_2$, $U^{(2)}$. The hyperparameters for training the NN objective functions and the NOM are the same as those shown in Table 1 and Table 2. The results are shown in Fig. 9. The red points are the results obtained using the NOM. The Non-dominated Sorting Genetic Algorithm (NSGA-II), a multiobjective optimization technique, is also applied for this problem. The results are shown in black points in Fig. 9. As can be seen from Fig. 9, the results obtained from both methods are almost identical.

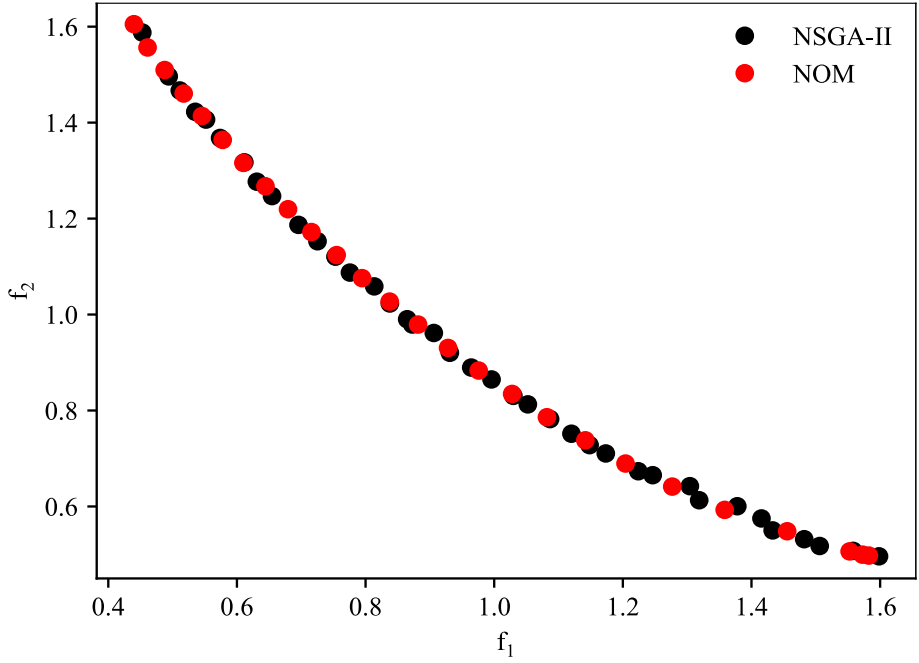

Fig. 9 Pareto optimum solutions.

### 3.3 Design of processing parameters in additive manufacturing

The developed NOM is flexible and not restricted to specific NN architectures or activation functions. In this section, we applied the NOM to design an engineering problem, i.e., the design of processing parameters in additive manufacturing (AM) to optimize the fatigue life of Ti-6Al-4V (Ti-64).

Ti-64 is a type of material extensively used in AM (Kumar & Ramamurty, 2020; Pegues, et al., 2020; Sandgren, et al., 2016; Sharma, et al., 2021; Wycisk, Emmelmann, Siddique, & Walther, 2013). Common applications of Ti-64 include turbine blades and compressors. In those cases, components may have fatigue failure due to the cyclic loadings (Chen & Liu, 2020b; Günther, et al., 2017). Therefore, the fatigue performance of AM Ti-64 is important in ensuring structural safety (Chen & Liu, 2021b). The processing parameters influence the AM quality for fatigue performance. The key factors are in- processing parameters and post-processing parameters. This paper aims to optimize the fatigue life by designing the processing parameters.

(Chen & Liu, 2021b) proposed a neural network approach for the fatigue modeling for AM Ti-64. The model is named Probabilistic Physics-guided Neural Network (PPgNN). That model can be used to obtain probabilistic stress-life (P-S-N) relationships (Chen, Liu, Zhang, & Liu, 2020) under different processing parameters. The PPgNN model is used as the NN objective function, as shown in the dashed line box in Fig. 10. The inputs of the PPgNN are fatigue parameters, AM in-processing parameters, and AM post- processing parameters. Specifically, fatigue parameters are stress amplitude $S$ and stress ratio $R$. Among all the AM in-processing parameters, scanning velocity $v$, laser power $P$, hatch distance $h$, and layer thickness $t$ are considered. Heat temperature $HT$ and heat time $Ht$ are the considered AM post-processing parameters. The outputs of the PPgNN are the statistics of the fatigue life, which are mean $\mu$ and standard deviation $\sigma$. The square neurons are designed for missing data problems. For a comprehensive introduction of the PPgNN, refer to (Chen & Liu, 2021b).

The architecture of the NOM shown in Fig. 10 is built according to Section 2.2.2 for constrained optimization. The loss function to be minimized by the NOM is

$$Loss(\text{NOM}) = -(\mu - 1.96\sigma) + \text{outputs of constraint neurons}, \qquad (22)$$

where $\mu - 1.96\sigma$ is the 2.5% lower bound of the fatigue life (Chen & Liu, 2020b). In other words, this problem is to maximize the lower bound. The constraints of this problem are the ranges of the input variables, and are listed next to the corresponding constraint neurons. The hyperparameters are the same as in previous examples. The stress amplitude and stress ratio are fixed to be 520 MPa and 0.1, respectively. The results of the NOM together with Nelder Mead, Genetic Algorithm, Differential Evolution, and Particle Swarm Optimization are shown in Table 4. The NOM achieves the same result of $\log_{10}(\mu - 1.96\sigma)$ as Genetic Algorithm, Differential Evolution, and Particle Swarm Optimization. The Nelder Mead provides a worse result in this case. This problem has more variables than those of the test examples shown in Section 3.1. As a result, the computational cost increases for all the other optimization algorithms. However, the computational cost is not shown to increase for the NOM.

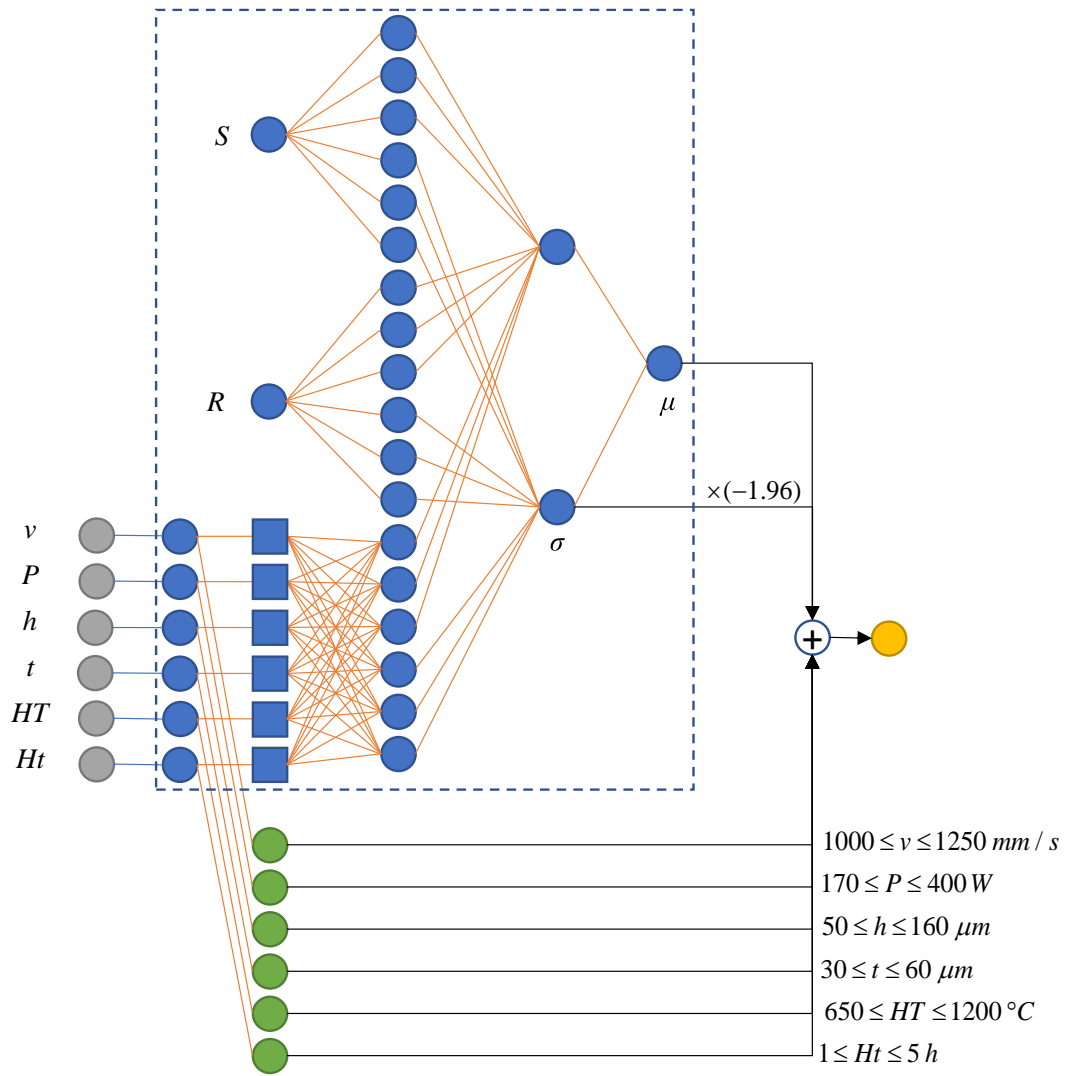

Fig. 10 Neural Optimization Machine for the design of processing parameters in additive manufacturing using Physics-guided Neural Network as the objective function.

Table 4 Results of different optimization algorithms for AM fatigue optimization.

|  | Nelder Mead | Genetic Algorithm | Differential Evolution | Particle Swarm Optimization | Neural Optimization Machine |
|---|---|---|---|---|---|
| $v$ (mm/s) | 1053.2 | 1067.2 | 1066.6 | 1064.1 | 1064.9 |
| $P$ (W) | 177.8 | 170.0 | 170.2 | 170.0 | 171.6 |
| $h$ (μm) | 86.3 | 90.2 | 88.9 | 90.6 | 90.8 |
| $t$ (μm) | 59.8 | 59.9 | 59.9 | 59.9 | 59.9 |
| $HT$ (°C) | 797.1 | 797.5 | 797.0 | 797.2 | 796.9 |
| $Ht$ (h) | 5.0 | 5.0 | 5.0 | 5.0 | 5.0 |
| $\log_{10}(\mu - 1.96\sigma)$ | 4.458 | 4.462 | 4.462 | 4.462 | 4.462 |
| Time/second | 32.2 | 105.3 | 261.6 | 162.2 | 21.6 |

## 4. Discussion

This section discusses the impact of initial weights and starting points. As stated at the end of Section 2.2.1, the NOM uses the strategy of random initial weights between the starting point layer and the input layer and multiple starting points to explore the global minimum. The impacts of these operations are discussed below.

During the training of the NOM, the weights and biases between the starting point layer and the input layer are updated. This is achieved by calculating the derivatives of the loss function of the NOM with respect to weights and biases according to the backpropagation algorithm:

$$dLoss/d\mathbf{b}^{(s)} = f_{\mathbf{b}}\left(\mathbf{b}^{(s)}, \mathbf{w}^{(s)}, \mathbf{X}_0\right), \tag{23}$$

and

$$dLoss/d\mathbf{w}^{(s)} = f_{\mathbf{w}}\left(\mathbf{b}^{(s)}, \mathbf{w}^{(s)}, \mathbf{X}_0\right), \tag{24}$$

respectively, where $Loss$ is the loss function of the NOM, $\mathbf{w}^{(s)}$ and $\mathbf{b}^{(s)}$ are the weights and biases between the starting point layer and the input layer, respectively, and $\mathbf{X}_0$ is the training data as well as the starting point of the NOM. Eqs. (23) and (24) show that the derivatives are the functions of the $\mathbf{w}^{(s)}$ and $\mathbf{b}^{(s)}$ at the previous step and $\mathbf{X}_0$. If the initial $\mathbf{w}^{(s)}$ and $\mathbf{b}^{(s)}$ are fixed to be zero and one, respectively, the calculations of the derivatives in Eqs. (23) and (24) at the first step are only the function of the starting point. In this way, the NOM will converge to the nearest local minimum from the starting point due to the steepest gradient descent algorithm. As stated in Section 2.2.1, before training the NOM, random initial values are assigned to $\mathbf{w}^{(s)}$. This introduces randomness in the first step of the gradient calculation. As the following updates of the weights and biases are functions of the results from the previous steps, the initial random weights are assigned to the NOM to find more local minima. Also, multiple starting points are used to train the NOM. The randomness of the initial weights and multiple starting points help the NOM to converge to multiple local minima, which is possible to obtain the global minimum.

The above discussion is demonstrated using the following optimization problem:

$$\begin{aligned} &\text{Minimize} \\ &f(x_1, x_2) = \frac{\sin^3(2x_1)\sin(2x_2)}{x_1^3(x_1 + x_2)} \\ &\text{subject to} \\ &0.1 \leq x_1 \leq 1 \\ &0.1 \leq x_2 \leq 7 \end{aligned} \tag{25}$$

This problem is modified from Problem 3 by deleting the first and second constraints. As shown in Fig. 7 (c), there are two local minima in the defined domain, and one of those is the global minimum. This problem aims to test whether the NOM can obtain both local minima or just one of those. The results of the number of local minima obtained by the NOM are shown in Table 5. Six cases are investigated according to the number of starting points and whether initial $\mathbf{w}^{(s)}$ are one or random. The initial $\mathbf{b}^{(s)}$ is zero for all cases. The test results in Table 5 show that the NOM can obtain both local minima with multiple starting points and random initial $\mathbf{w}^{(s)}$. Therefore, we use random initial $\mathbf{w}^{(s)}$ and 5 starting points for all the previous optimization problems. The starting points correspond to 5 smallest objective function values determined by the grid search as stated in Section 2.2.1. This strategy benefits finding multiple local minima, which is possible to find the global minima. However, the other optimization techniques used in the previous sections only try to find the global minima.

Table 5 Number of local minima obtained by the NOM.

|  | 1 starting point | | Multiple starting points |
|---|---|---|---|
|  | Training NOM once | Training NOM multiple times |  |
| Initial $\mathbf{w}^{(s)}$ is one. | 1 | 1 | 2 |
| Initial $\mathbf{w}^{(s)}$ is random. | 1 | 2 | 2 |

## 5. Conclusions

Neural Optimization Machine (NOM), a novel neural network approach, is proposed for constrained optimization of neural network models. The objective functions for the NOM are NN models. The optimization process is conducted by the neural network's built-in backpropagation algorithm. The NOM solves optimization problems by extending the architecture of the NN objective function model. This is achieved through the appropriate design of the NOM's structure, activation function, and loss function. The NOM is very flexible and is extended for multiobjective optimization. The NOM is tested using numerical optimization problems for constrained optimization and multiobjective optimization. The results obtained from the NOM are compared with the Nelder Mead, Genetic Algorithm, Differential Evolution, and Particle Swarm Optimization for single-objective optimization, and Non-dominated Sorting Genetic Algorithm (NSGA-II) for multiobjective optimization. The NOM is then applied for the design of processing parameters in additive manufacturing. Based on the investigation of this paper, the following are the conclusions.

1. The NN objective function can have arbitrary architectures and activation functions.
2. The application of the NOM is not limited to specific optimization problems, e.g., linear and quadratic programming.
3. Multiple local minima can be found, which provides the potential for finding the global minimum.
4. The increase of dimension of design variables does not increase the computational cost significantly for the NOM.

## Acknowledgments

The research in this paper was partially supported by funds from NASA University Leadership Initiative program (Contract No. NNX17AJ86A, Project Officer: Dr. Anupa Bajwa, Principal Investigator: Dr. Yongming Liu). The support is gratefully acknowledged.